\documentclass[sigconf]{acmart}
\renewcommand\footnotetextcopyrightpermission[1]{}
\AtBeginDocument{%
  }

\usepackage{float}
\usepackage{subcaption}  
\usepackage{siunitx}

\begin{document}
\title{HoverAI: An Embodied Aerial Agent for Natural Human-Drone Interaction}

\author{Yuhua Jin}
\authornote{These authors contributed equally to this work.}
\affiliation{%
  \institution{Chinese University of Hong Kong, Shenzhen}
  \city{}
  \state{Guangdong}
  \country{China}}
\email{yuhuajin@cuhk.edu.cn}

\author{Nikita Kuzmin}
\authornotemark[1]
\affiliation{%
  \institution{Skolkovo Institute of Science and Technology}
  \city{Moscow}
  \country{Russia}}
\email{Nikita.Kuzmin@skoltech.ru}

\author{Georgii Demianchuk}
\authornote{These authors contributed equally to this work.}
\affiliation{%
  \institution{Skolkovo Institute of Science and Technology}
  \city{Moscow}
  \country{Russia}}
\email{Georgii.Demianchuk@skoltech.ru}

\author{Mariya Lezina}
\authornotemark[2]
\affiliation{%
  \institution{Skolkovo Institute of Science and Technology}
  \city{Moscow}
  \country{Russia}}
\email{Mariya.Lezina@skoltech.ru}

\author{Fawad Mehboob}
\affiliation{%
  \institution{Skolkovo Institute of Science and Technology}
  \city{Moscow}
  \country{Russia}}
\email{Fawad.Mehboob@skoltech.ru}

\author{Issatay Tokmurziyev}
\affiliation{%
  \institution{Skolkovo Institute of Science and Technology}
  \city{Moscow}
  \country{Russia}}
\email{issatay.tokmurziyev@skoltech.ru}

\author{Miguel Altamirano Cabrera}
\affiliation{%
  \institution{Skolkovo Institute of Science and Technology}
  \city{Moscow}
  \country{Russia}}
\email{m.altamirano@skoltech.ru}

\author{Muhammad Ahsan Mustafa}
\affiliation{%
  \institution{Skolkovo Institute of Science and Technology}
  \city{Moscow}
  \country{Russia}}
\email{Ahsan.Mustafa@skoltech.ru}

\author{Dzmitry Tsetserukou}
\affiliation{%
  \institution{Skolkovo Institute of Science and Technology}
  \city{Moscow}
  \country{Russia}}
\email{d.tsetserukou@skoltech.ru}

\renewcommand{\shortauthors}{Jin et al.}

\begin{abstract}
Drones operating in human-occupied spaces suffer from insufficient communication mechanisms that create uncertainty about their intentions. We present HoverAI, an embodied aerial agent that integrates drone mobility, infrastructure-independent visual projection, and real-time conversational AI into a unified platform. Equipped with a MEMS laser projector, onboard semi-rigid screen, and RGB camera, HoverAI perceives users through vision and voice, responding via lip-synced avatars that adapt appearance to user demographics. The system employs a multimodal pipeline combining VAD, ASR (Whisper), LLM-based intent classification, RAG for dialogue, face analysis for personalization, and voice synthesis (XTTS v2). Evaluation demonstrates high accuracy in command recognition (F1: 0.90), demographic estimation (gender F1: 0.89, age MAE: 5.14 years), and speech transcription (WER: 0.181). By uniting aerial robotics with adaptive conversational AI and self-contained visual output, HoverAI introduces a new class of spatially-aware, socially responsive embodied agents for applications in guidance, assistance, and human-centered interaction.
\end{abstract}


\begin{teaserfigure}
    \centering
    \includegraphics[width=0.8\textwidth]{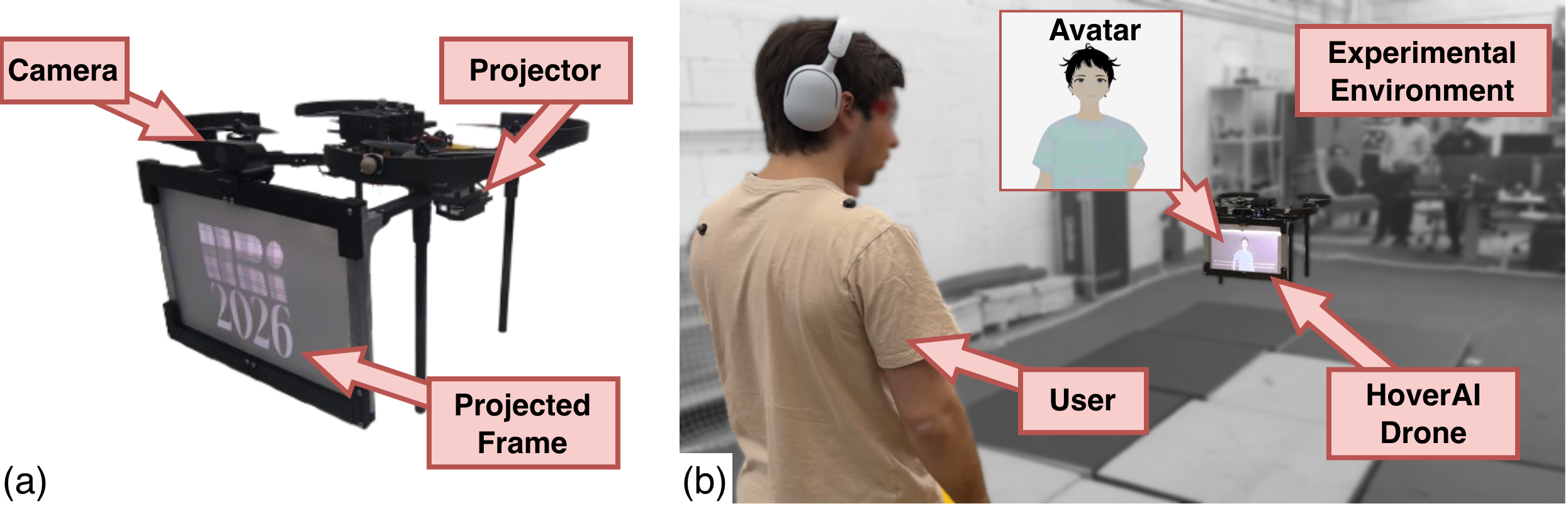} 
    \caption{The HoverAI interactive aerial interface: (a) The HoverAI drone featuring an RGB camera for user perception and a front-mounted semi-rigid projection screen for displaying the interactive avatar; (b) User interacting with HoverAI in an experimental environment, wearing headphones for audio input/output while the drone hovers and displays a projected avatar.}
    \Description{The figure illustrates the HoverAI system in two parts: (a) shows a drone with labeled components including a camera, projector, and projected frame; (b) depicts a user interacting with the HoverAI drone in an experimental environment, where an avatar is displayed on the drone’s screen, visually connecting the user and the drone in a shared physical space.}
    \label{fig:teaser}
\end{teaserfigure}


\maketitle
\section{Introduction}
Digital interfaces are increasingly moving beyond fixed screens into physical environments, reshaping how people interact with computational systems. Embodied AI agents offer a promising paradigm for more natural and spatially-aware human-computer interaction ~\cite{embodiedAI2025, spatialXR2024}. As spatial computing evolves, such agents can make digital information accessible exactly where it is needed, enabling seamless integration of content into shared environments. However, challenges in mobility, perception, communication, and social presence must be addressed to fully realize this vision.

Traditional interfaces such as smartphones and monitors confine users to fixed locations and require explicit attention, limiting their suitability for dynamic, multi-user settings. Augmented and virtual reality systems provide immersive experiences but often isolate users behind head-mounted displays and depend on controlled environments. These limitations motivate the development of mobile platforms that can interact with people directly within everyday physical spaces.

Ground robots are a common solution, yet their dependence on floor-based navigation restricts mobility. They must avoid obstacles, navigate through doorways, and cope with constantly changing layouts, making them unreliable in crowded or dynamic environments such as airports or museums. Drones, in contrast, provide unrestricted three-dimensional mobility: they can hover at eye level, bypass obstacles from above, and accompany users without obstructing pedestrian flow. These properties make drones attractive as interactive embodied agents. Recent work has demonstrated the feasibility of integrating advanced AI models onboard UAVs for real-time reasoning \cite{lykov2025cognitivedrone}, but natural human-drone communication remains largely unresolved. Studies in human-drone interaction show that uncertainty about drone intentions and state significantly reduces trust and usability in public spaces \cite{dronePublicSpaces2025, lingam2025challenges}.

Prior efforts to create drone-based visual interfaces illustrate both the potential and limitations of current approaches. Large-scale drone light shows enable impressive aerial displays but are designed for passive viewing rather than close interaction \cite{RSS}. Systems such as BitDrones attached small displays to drones, yet these solutions are constrained by weight, power, and viewing-angle limitations \cite{BitDrones}. Projection-based approaches like LightAir and MaskBot enable expressive visuals, but rely on external surfaces or infrastructure, preventing fully mobile interaction \cite{LightAir, cabrera2020maskbot}. Flying Light Specks achieved detailed visualizations using coordinated drones and fixed projectors, but similarly requires pre-installed equipment \cite{LightSpecks}. Recent work on expressive drone avatars explored social presence through digital faces \cite{FacialEmotions}, yet lacked real-time conversational intelligence and infrastructure-independent output. To date, no aerial system has combined onboard conversational AI, adaptive visual projection, and real-time social interaction in a single lightweight platform.

We present HoverAI, an interactive aerial agent that combines drone mobility, ultra-light MEMS laser projection, real-time speech understanding, and adaptive avatar expression in a self-contained system. Using an onboard RGB camera and projection screen, HoverAI perceives users and communicates via voice and projected imagery without relying on external infrastructure. A multimodal pipeline (VAD, ASR, lightweight LLMs, and RAG) enables contextual dialogue, while real-time face analysis personalizes a lip-synced avatar. This closed-loop design allows HoverAI to function as a socially present, spatially aware embodied agent capable of natural interaction in shared environments.

\section{System Architecture}

\begin{figure}[h]
  \centering
  \includegraphics[width=\linewidth]{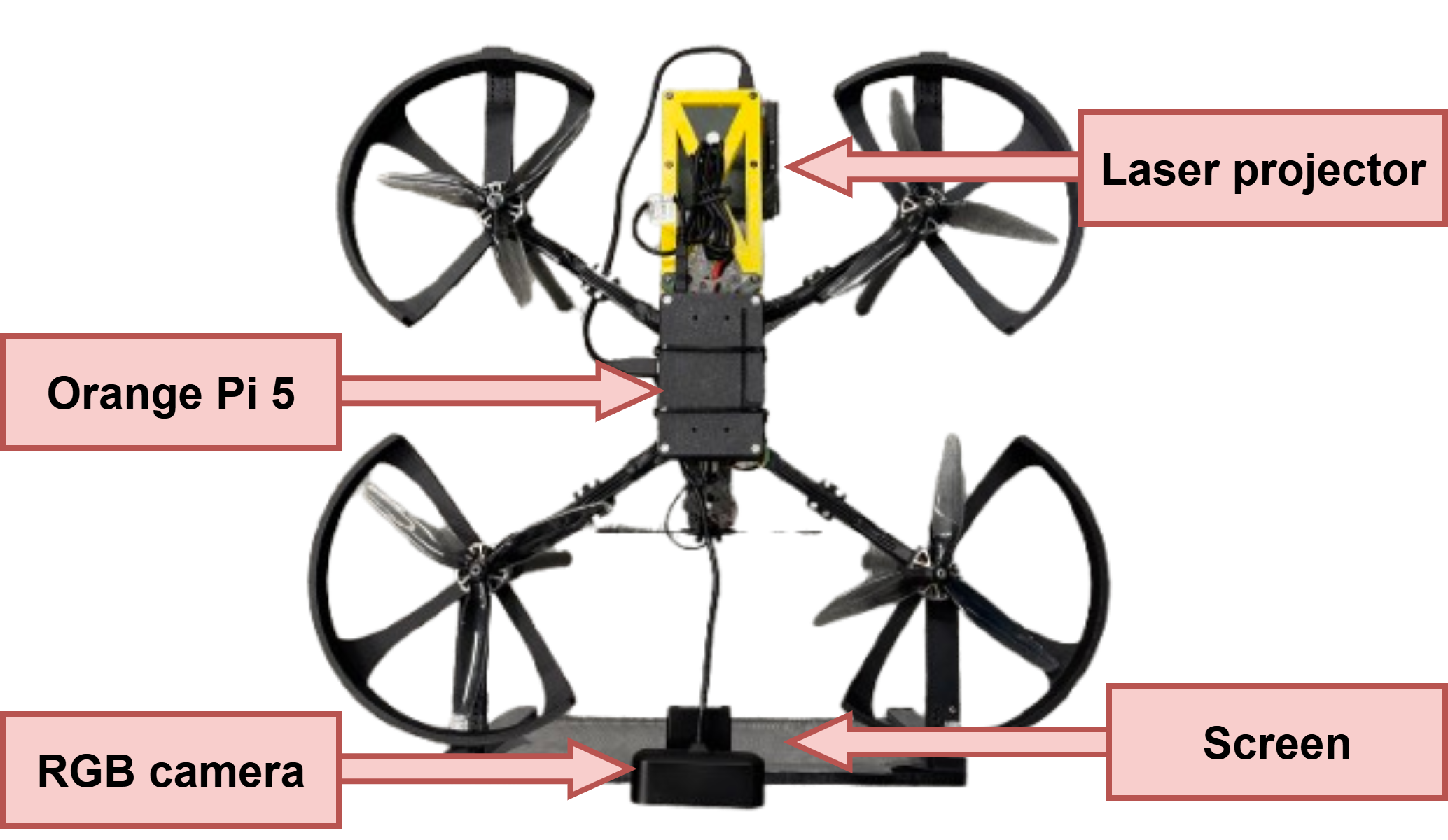}
  \caption{HoverAI quadcopter key hardware components.} 
  \label{fig:drone}
  \Description{This figure shows the key components of the drone: MEMS-based laser projector with a front-mounted flexible projection film, camera, and Orange Pi 5.}
\end{figure}

Currently, no aerial system combines onboard conversational AI, environment-independent visual output, and real-time social adaptation in a single lightweight platform. HoverAI addresses this gap through a self-contained quadrotor that integrates flight, visual projection, environmental perception, and conversational interaction. As shown in Fig.~\ref{fig:drone}, the 1.2 kg platform comprises: an Orange Pi 5 single-board computer for real-time processing, a front-facing RGB camera (1080p, 30 fps), and a MEMS laser projector (720p, 30 fps, 85 g) paired with a semi-rigid projection film, enabling infrastructure-free visual display during flight.

\subsection{Hardware Architecture}

The Laser Scanning Projection (LSP) module features a 2D MEMS scanning mirror delivering 720p resolution without the weight or power constraints of conventional displays. A semi-rigid polycarbonate film (0.3 mm, 40 g) suspended 15 cm in front of the projector remains stable under airflow and vibrations, projecting clear visuals without requiring external surfaces. The Orange Pi 5 manages flight control via a  Speedybee F405V4 flight controller and communicates wirelessly with a ground station PC over WiFi (5 GHz, ~50 ms latency). User speech is captured via a close-talking headphone microphone to maximize signal-to-noise ratio in indoor environments.

\subsection{Interaction Pipeline}

\begin{figure}[t!]
  \centering
  \includegraphics[width=\columnwidth]{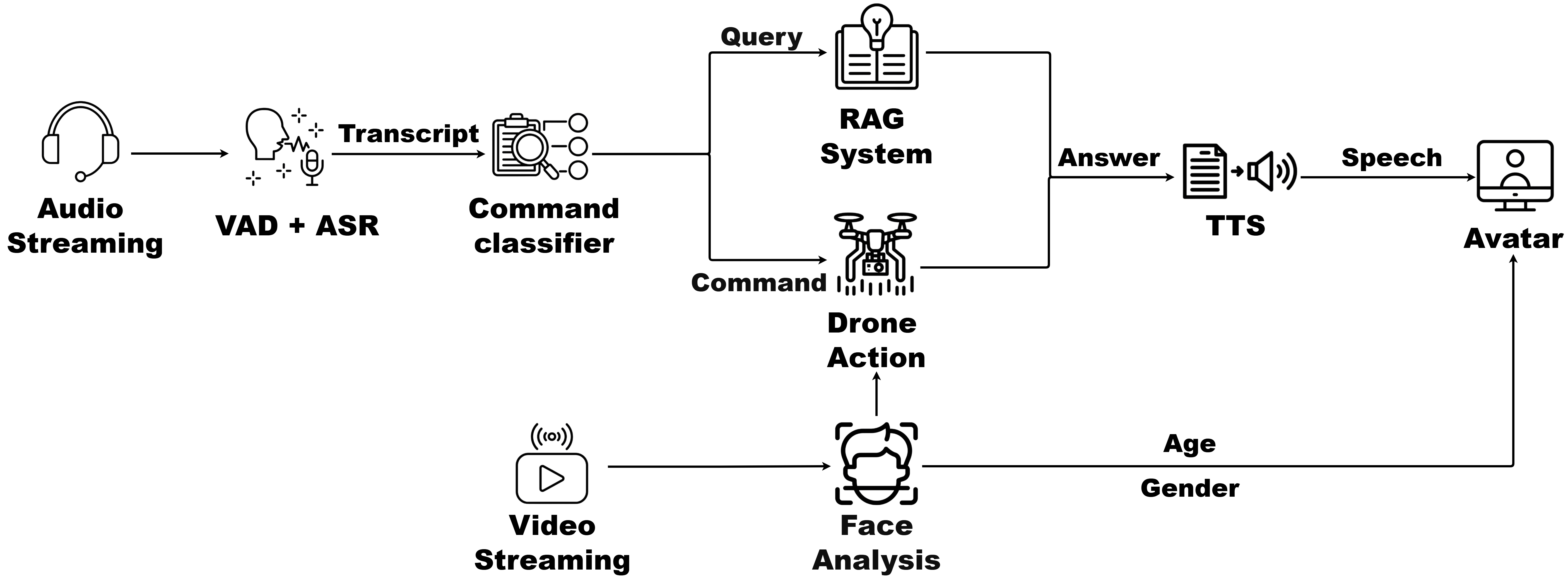}
 \caption{HoverAI pipeline with audio analysis, video-based face processing, and TTS-driven avatar output.}

  \Description{This figure shows the system architecture diagram showing data flow from user input to drone output. On the left, audio input goes through VAD and ASR to transcript, then to the command classifier. Video input goes to face analysis for age and gender. Command classifier sends either a command to drone action or a query to the RAG system. RAG returns an answer to TTS, which generates speech. Speech and face analysis data feed into the avatar display. Drone action controls physical movement. All components are connected with arrows indicating the direction of data flow.}
  \label{fig:system-arch}
\end{figure}

As illustrated in Fig.~\ref{fig:system-arch}, HoverAI processes two parallel input streams. Audio is transmitted via WiFi to the ground station, where it undergoes spectral noise reduction, Voice Activity Detection (VAD) \cite{Silero}, and Automatic Speech Recognition using Whisper-medium.en \cite{radford2023robust} (average WER: 0.181). Transcripts are classified by gemma:7b-instruct \cite{gemma7binstruct2024} into:

\begin{itemize}
    \item \textbf{Structured commands} (``follow,'' ``land,'' ``stay,'' ``explore face'') forwarded to the drone control module for immediate execution.
    \item \textbf{Conversational queries} routed to a Retrieval-Augmented Generation (RAG) system with a domain-specific knowledge base (museum artefacts, FAQ), ensuring grounded responses and minimizing hallucination.
\end{itemize}

Generated responses are synthesized via XTTS v2 Text-to-Speech, which adapts voice characteristics (pitch, timbre) to match the selected avatar demographic \cite{casanova2024xtts}.

Concurrently, the video stream undergoes face analysis via InsightFace \cite{insightface2017}, estimating age and gender to select among four predefined avatars: young woman ($< 30$), adult woman ($\geq 30$), young man ($< 30$), and adult man ($\geq 30$). When no face is detected, a gender-neutral default avatar is displayed.

The synthesized speech plays through user's headphones while a lip-synced avatar with subtitles renders on the drone screen at $\sim25$ fps (total pipeline latency: 800--1200 ms). This closed-loop architecture enables HoverAI to function as a socially present, embodied agent that adapts appearance, voice, and behavior to user input, supporting natural turn-taking and co-presence.

The modular design supports future extensions, including SLAM-based autonomous navigation, multi-drone swarm coordination for large-scale displays, and expanded knowledge bases with real-time web querying while maintaining response reliability.

\section{Evaluation}

To validate HoverAI's multimodal interaction capabilities, we conducted a benchmark evaluation measuring performance across speech recognition, intent classification, and demographic estimation.

\subsection{Experimental Setup}

We recruited 12 participants (6 male, 6 female, aged 22-48) to interact with HoverAI in a controlled indoor laboratory environment (6$\times$6 m). Each participant completed a 5-minute interaction session involving:
\begin{itemize}
    \item \textbf{Speech tasks}: 20 conversational queries (general knowledge, navigation) and 10 structured commands (``follow me,'' ``land,'' ``stay,'' ``explore face'')
    \item \textbf{Vision tasks}: Continuous face tracking during interaction for demographic estimation
\end{itemize}

Speech was captured via close-talking headphones in ambient noise conditions (45-50 dB). All sessions were recorded with informed consent. The RAG knowledge base contained 150 curated facts about robotics and museum artifacts.

\subsection{Results}

\begin{figure}[h]
\centering
\includegraphics[width=\linewidth]{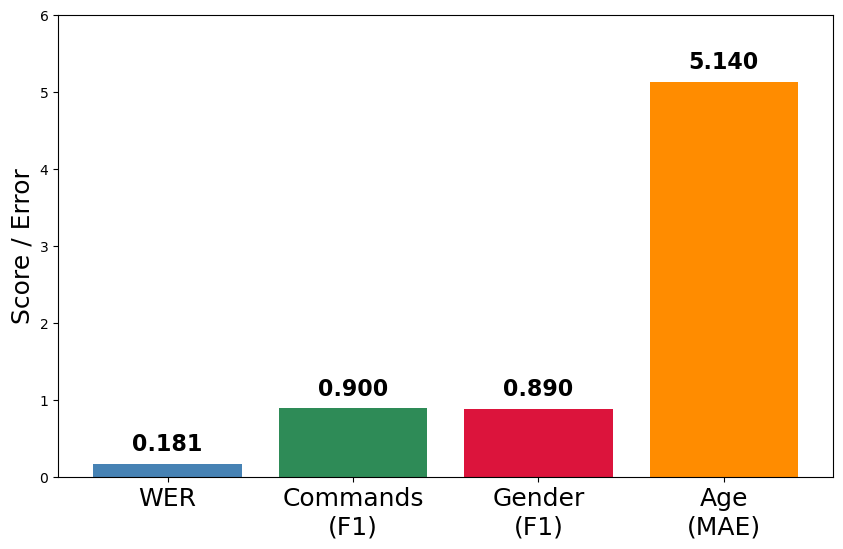}
\caption{Performance metrics averaged across 12 participants: speech transcription (WER: 0.181), command recognition (F1: 0.90), gender classification (F1: 0.89), and age estimation (MAE: 5.14 years).}
\Description{Bar chart showing four performance metrics: WER of 0.181, Command F1 of 0.9, Gender F1 of 0.89, and Age MAE of 5.14 years.}
\label{fig:metrics}
\end{figure}

As shown in Fig.~\ref{fig:metrics}, HoverAI achieved strong performance across all modalities:

\begin{itemize}
    \item \textbf{Speech Transcription (WER: 0.181)}: Whisper-medium.en demonstrated reliable ASR despite ambient noise, with most errors occurring on technical terminology.
    \item \textbf{Command Recognition (F1: 0.90)}: The gemma:7b classifier correctly distinguished commands from queries in 90\% of cases, with confusion primarily between ``stay'' and conversational ``wait'' statements.
    \item \textbf{Gender Estimation (F1: 0.89)}: InsightFace achieved robust classification across lighting conditions and viewing angles ($\pm$ \ang{30}).

    \item \textbf{Age Estimation (MAE: 5.14 years)}: While absolute age error averaged 5.14 years, binary classification ($< 30$ vs. $\geq30$) for avatar selection achieved 91.7\% accuracy, sufficient for demographic adaptation.
\end{itemize}

End-to-end pipeline latency averaged 950 ms ($\pm$120 ms) from speech onset to avatar response, supporting natural conversational turn-taking. No system crashes occurred during 60 minutes of total interaction time.

\section{Applications and Use Cases}

HoverAI's combination of mobility, conversational AI, and adaptive visual presence enables several application scenarios:

\textbf{Museum and Educational Guidance}: HoverAI can follow visitors through exhibitions, projecting contextual narratives, multilingual subtitles, or 3D reconstructions directly aligned with physical artefacts. Its spatial mobility enables optimal positioning for visibility, offering an alternative to static displays or audio guides for users with limited mobility.

\textbf{Assistive Communication}: For users with motor disabilities or speech impairments, HoverAI can display speech-to-text output, mirror smartphone content, or serve as a visual proxy during remote calls. Its hands-free operation and human-scale presence are valuable in home healthcare and eldercare contexts where conventional devices may be inaccessible.

\textbf{Personal Companion}: HoverAI can provide ambient assistance in everyday environments, hovering at eye level to deliver contextual information or social engagement through natural conversation, creating a persistent visual presence without requiring touch or wearable devices.

\section{Discussion}

\subsection{Limitations}

The system demonstrates robust technical performance, achieving high accuracy in speech recognition, command classification, and demographic estimation, and establishes a compelling foundation for socially responsive human-drone interaction.

However, several technical and methodological limitations constrain the current prototype. Flight time is limited to $\sim12$ minutes by battery capacity, restricting deployment duration. Stable projection requires indoor operation with minimal wind ($<$0.5 m/s) and controlled lighting; outdoor use degrades visibility. WiFi-based audio streaming limits operational range to $\sim15$ m. The semi-rigid screen occasionally exhibits vibration artefacts during aggressive maneuvers, requiring conservative flight profiles.

Critically, our evaluation focused on technical performance metrics rather than user perception. While we demonstrated reliable speech recognition and demographic estimation, we did not assess whether the adaptive avatar actually reduces uncertainty or enhances social presence. 
Future work should therefore include controlled user studies measuring perceived safety, trust, and clarity of drone intentions during interaction. 
Qualitative feedback and comparative experiments with non-adaptive or non-visual drone interfaces would help determine how much the projected avatar contributes to usability and overall user experience.

Deploying a drone that adapts its appearance based on users' age and gender and displays a human-like avatar raises important ethical questions. Recording people’s faces and voices in public, especially without their consent, can violate privacy. Mistakes in estimating age or gender may lead to misrepresentation or bias, and the lifelike avatar might make people think the drone has intentions it does not actually have. To address these risks, future work should include clear indicators when recording is active, obtain user consent, handle data transparently, and comply with privacy laws.

The RAG system is limited to pre-defined knowledge bases ($\sim150$ facts) and cannot retrieve real-time information or handle out-of-domain queries gracefully, occasionally producing generic responses.

\subsection{Future Directions}

Near-term improvements include SLAM-based autonomous navigation for spatial positioning, extended battery life through optimized power management, and expanded knowledge bases with vetted external source integration. 

Robustness to real-world acoustic environments is another critical direction: our current evaluation was conducted in controlled indoor settings with moderate ambient noise, but practical deployments in museums, educational spaces, or urban outdoor areas often involve higher noise levels and overlapping speech. Future work should therefore evaluate and enhance the system’s speech perception pipeline under diverse acoustic conditions, potentially incorporating noise-robust ASR models to maintain performance.

Longer-term research directions include multi-drone swarm coordination for large-scale collaborative displays, 3D volumetric visualization, and distributed perception across coordinated units. Investigating outdoor deployment with brighter projection and stabilized screen mechanisms would broaden applicability.

\section{Conclusion}

We presented HoverAI, an embodied aerial agent that integrates infrastructure-independent visual projection, real-time conversational AI, and demographic-adaptive avatar generation into a self-contained mobile platform. By combining MEMS laser projection with a semi-rigid screen, multimodal perception through vision and speech, and closed-loop interaction via LLM-based dialogue and face analysis, HoverAI demonstrates a new approach to spatially-aware human-drone interaction. Evaluation across 12 participants showed robust performance in speech recognition (WER: 0.181), command classification (F1: 0.90), and demographic estimation, establishing technical feasibility for applications in guidance, assistance, and companionship. HoverAI represents a step toward mobile, socially responsive interfaces that bring digital content into shared physical spaces in more human-centered ways.

\section*{Acknowledgements} 
Research reported in this publication was financially supported by the RSF grant No. 24-41-02039.




\end{document}